\documentclass[preprint,1p,11pt]{ISAS_IR}

\usepackage[utf8]{inputenc}
\usepackage[T1]{fontenc}
\usepackage[per-mode=symbol]{siunitx} 
\usepackage{subfig}
\usepackage{nicefrac} % for fractions with slash: a/b
\usepackage{comment} % block comment: \begin{comment} ... \end{comment}

\usepackage{microtype} % Optischer Randausgleich

\usepackage{amsmath,amssymb,amsfonts}
\usepackage[ruled]{algorithm2e} % ,vlined
\usepackage{etoolbox}
\makeatletter
\patchcmd{\@algocf@start}% <cmd>
{-1.5em}% <search>
{0pt}% <replace>
{}{}% <success><failure>
\makeatother

\usepackage{graphicx}
\usepackage{url}

\usepackage{mathtools}
\mathtoolsset{showonlyrefs}

\allowdisplaybreaks

\usepackage{acro}

\DeclareAcronym{TOA}		{short={TOA},		    long={time of arrival}, long-plural-form={times of arrival}}
\DeclareAcronym{TDOA}		{short={TDOA},		    long={time difference of arrival}, long-plural-form={time differences of arrival}}
\DeclareAcronym{DOTA}		{short={DOTA},		    long={difference of times of arrival}, long-plural-form={differences of times of arrival}, short-plural-form={DOTAs}}
\DeclareAcronym{ROTA}		{short={ROTA},		    long={round trip time of arrival}, long-plural-form={round trip times of arrival}}
\DeclareAcronym{LTDOA}		{short={L\hbox{-}TDOA},	long={local time difference of arrival}, long-plural-form={local time differences of arrival}}
\DeclareAcronym{LDOTA}		{short={L\hbox{-}DOTA},	long={local differences of times of arrival}, long-plural-form={local differences of times of arrival}}
\DeclareAcronym{DOTD}		{short={DOTD},		    long={difference of time differences}}
\DeclareAcronym{TTT}		{short={TTT},		    long={target transmission time}}
\DeclareAcronym{TDOT}		{short={TDOT},		    long={time difference of transmissions}, long-plural-form={time difference of transmissions}}
\DeclareAcronym{SSR}		{short={SSR},	    	long={secondary surveillance radar}}
\DeclareAcronym{PSR}		{short={PSR},	    	long={primary surveillance radar}}
\DeclareAcronym{LORAN}		{short={LORAN},	    	long={long range navigation system}}
\DeclareAcronym{MLAT}		{short={MLAT},	    	long={multilateration}}
\DeclareAcronym{MLE}		{short={MLE},	    	long={maximum likelihood estimator}}
\DeclareAcronym{RSS}		{short={RSS},	    	long={received signal strength}}
\DeclareAcronym{AOA}		{short={AOA},	    	long={angle of arrival}}
\DeclareAcronym{ML}		    {short={ML},	    	long={maximum likelihood}}
\DeclareAcronym{MAP}	    {short={MAP},	    	long={maximum a posteriori}}
\DeclareAcronym{MMSE}		{short={MMSE},	    	long={minimum mean square error}}
\DeclareAcronym{RMSE}		{short={RMSE},	    	long={root mean square error}}
\DeclareAcronym{LS}		    {short={LS},	    	long={least squares}}
\DeclareAcronym{LM}		    {short={LM},	    	long={Levenberg-Marquardt}}
\DeclareAcronym{GNSS}		{short={GNSS},	    	long={global navigation satellite systems}}
\DeclareAcronym{GPS}		{short={GPS},	    	long={Global Positioning System}}
\DeclareAcronym{ADSB}		{short={ADS-B},	    	long={automatic dependent surveillance -- broadcast}}
\DeclareAcronym{TOF}		{short={TOF},	    	long={time of flight}}
\DeclareAcronym{LORAWAN}	{short={LoRaWAN},	  	long={Long Range Wide Area Network}}
\DeclareAcronym{3DUSCT}	    {short={3D-USCT},	  	long={3D-Ultrasound Computer Tomography}}
\DeclareAcronym{EKF}		{short={EKF},	    	long={Extended Kalman Filter}}
\DeclareAcronym{UKF}		{short={UKF},	    	long={Unscented Kalman Filter}}
\DeclareAcronym{SDR}		{short={SDR},	    	long={software--defined radio}}
\DeclareAcronym{CPU}		{short={CPU},	    	long={central processing unit}}
\DeclareAcronym{CRLB}		{short={CRLB},	    	long={Cramér--Rao lower bound}}
\DeclareAcronym{VR}		    {short={VR},	    	long={virtual reality}}
\DeclareAcronym{SME}		{short={SME},	    	long={Symmetric Measurement Equation}}
\DeclareAcronym{LCD}		{short={LCD},	    	long={Localized Cumulative Distribution}}
\DeclareAcronym{EM}	    	{short={EM},	    	long={expectation--maximization}}
\DeclareAcronym{S2KF}	    {short={S$^2$KF},	    long={Smart Sampling Kalman Filter}}
\DeclareAcronym{ODE}	    {short={ODE},	    	long={ordinary differential equation}}
\DeclareAcronym{GM}	        {short={GM},	    	long={Gaussian mixture}}
\DeclareAcronym{DM}	        {short={DM},	    	long={Dirac mixture}}
\DeclareAcronym{LRKF}	    {short={LRKF},	    	long={linear regression Kalman filter}}
\DeclareAcronym{CoDiCo}	    {short={CoDiCo},	    long={coupled discrete and continuous densities}} 
\DeclareAcronym{vMF}		{short={vMF},	    	long={von Mises--Fisher}}
\DeclareAcronym{DMD}		{short={DMD},	    	long={Dirac mixture density}}
\DeclareAcronym{PDF}		{short={PDF},	    	long={probability density function}}
\DeclareAcronym{CDF}		{short={CDF},	    	long={cumulative density function}}
\DeclareAcronym{PCD}		{short={PCD},	    	long={projected cumulative distribution}}

\def\vec#1{\underline{#1}}

\def\mat#1{{\mathbf #1}}

\def\1_2{{\frac{1}{2}}}

\def\dd{{\,\operatorfont{d}}} % "d" operator for integration 
\def\T{ ^\top } % Transpose 
\def\op#1{{\operatorfont{#1}}}

\def\NewR{\mathbb{R}} % {{\rm I\hspace{-.17em}R}}
\def\Gauss{{\cal N}}

\def\Sec#1{Sec.~\ref{#1}}

\def\Fig#1{Fig.~\ref{#1}}

\usepackage{color}

\newcommand\paren[1]{\left( #1 \right)}              % \paren{a}     (a)   (normal parentheses)
\newcommand\brackets[1]{\left[ #1 \right]}           % \brackets{a}  [a]   (square brackets)
\newcommand\braces[1]{\left\lbrace #1 \right\rbrace} % \braces{a}    {a}
\newcommand\abs[1]{\left| #1 \right|}                % \abs{a}       |a|
\DeclareMathOperator*{\argmax}{arg\,max}

\def\fd#1#2{f_{{#1}}^{#2\!}}             % \fd{k}{v}(v) apperas as f_k^v(v) 

\SetKw{Fun}{Function} 
\SetKwFunction{KwSamplePrior}{samplePrior} 
\SetKwFunction{KwRound}{round} 
\SetKwFunction{KwGetStdNormalSamples}{stdNormalSamples} 
\SetKwFunction{KwChol}{chol} 
\SetKwFunction{KwNormalizeLOne}{normalizeL1} 
\SetKwFunction{KwMin}{min} 

\SetKwFunction{KwSampleProjected}{sample1D} 
\SetKwFunction{KwSampleSTwo}{sampleS1} 
\SetKwFunction{KwCumtrapz}{cumtrapz} 
\SetKwFunction{KwFindAdjacent}{adjacent} 
\SetKwFunction{KwRoots}{roots} 
\SetKwFunction{KwSelectBest}{select\_best} 
\SetKwFunction{KwSort}{sort} 
\SetKwFunction{KwProjectOrth}{project\_orth} 
\SetKwFunction{KwProject}{project} 
\SetKwFunction{KwRand}{rand} 
\SetKwFunction{KwAngToSTwo}{ang2S2} 
\SetKwFunction{KwAtanTwo}{atan2} 
\SetKwFunction{KwBackproject}{backproject}

\def\param{ {\mat \Theta} } % \param   parameter vector 
\def\obsrv{ {\mat Y} }      % \obsrv   observed data vector 
\def\hidden{ {\mat H} }     % \hidden  hidden data vector 

\def\hiddenStep#1{ {\widehat\hidden^{(#1)}} }     % \hiddenStep{r}  hidden data vector, recursion index r 
\def\paramStep#1{ {\widehat\param^{(#1)}} }      % \paramStep{r}  parameter vector, recursion index r 

\def\w{ {w} } % weighting factor 
\def\mean{ {\vec\mu} } % mean vector 
\def\C{ { \mat C } } % covariance matrix 

\def\s { {\vec s} }
\def\ws{ {\alpha} } % weighting factor 

\def\x { {\vec x} } 
 
\begin{document}

\begin{frontmatter}

\title{Gaussian Mixture Estimation \\from Weighted Samples}

\author{Uwe~D.~Hanebeck}

\author{Daniel~Frisch}

\address{Intelligent Sensor-Actuator-Systems Laboratory (ISAS)\\
	Institute for Anthropomatics and Robotics\\
	Karlsruhe Institute of Technology (KIT), Germany\\
	e-mail: {\normalfont \texttt{daniel.frisch@ieee.org, uwe.hanebeck@ieee.org}}}

\begin{abstract}
%
% Best fit of GM to samples
%
We consider estimating the parameters of a Gaussian mixture density with a given number of components best representing a given set of weighted samples. 
%
% Adopt density interpretation
%
We adopt a density interpretation of the samples by viewing them as a discrete Dirac mixture density over a continuous domain with weighted components.
%
% Density re-approximation
%
Hence, Gaussian mixture fitting is viewed as density re-approximation.
%
% Speed-up: EM
%
In order to speed up computation, an expectation--maximization method is proposed that properly considers not only the sample locations, but also the corresponding weights.
%
% Method from literature not correct
%
It is shown that methods from literature do not treat the weights correctly, resulting in wrong estimates.
%
% Scalar counterexamples
%
This is demonstrated with simple counterexamples.
%
% but method works in all dimensions
%
The proposed method works in any number of dimensions with the same computational load as standard Gaussian mixture estimators for unweighted samples.
\end{abstract}

\end{frontmatter}

%TODO 2D Example Picture(s)  

\section{Introduction}

%
% GM estimation standard problem
%
Gaussian mixture (GM) \acuse{GM} estimation is ubiquitous in signal processing and machine learning.
%
% Samples given, GM parameters desired
%
Given a set of samples, the parameters of a \ac{GM} are determined in such a way as to best fit the samples in a maximum likelihood way.
%
% Many methods available, EM the most prevalent 
%
Solutions for equally weighted samples are readily available, \ac{EM} based methods being the most prevalent because of low computational requirements and ease of implementation.

\begin{figure}[!ht] 
    \begin{center} 
        \subfloat[``separated'' components] {\includegraphics[width=.371\linewidth] {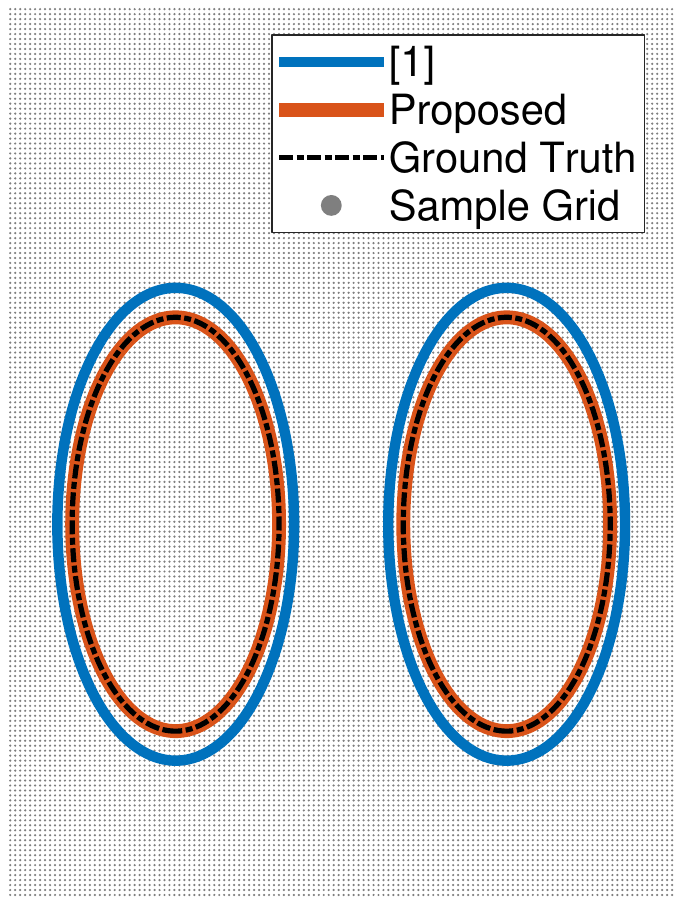}} 
        \subfloat[``overlapping'' components] {\includegraphics[width=.322\linewidth] {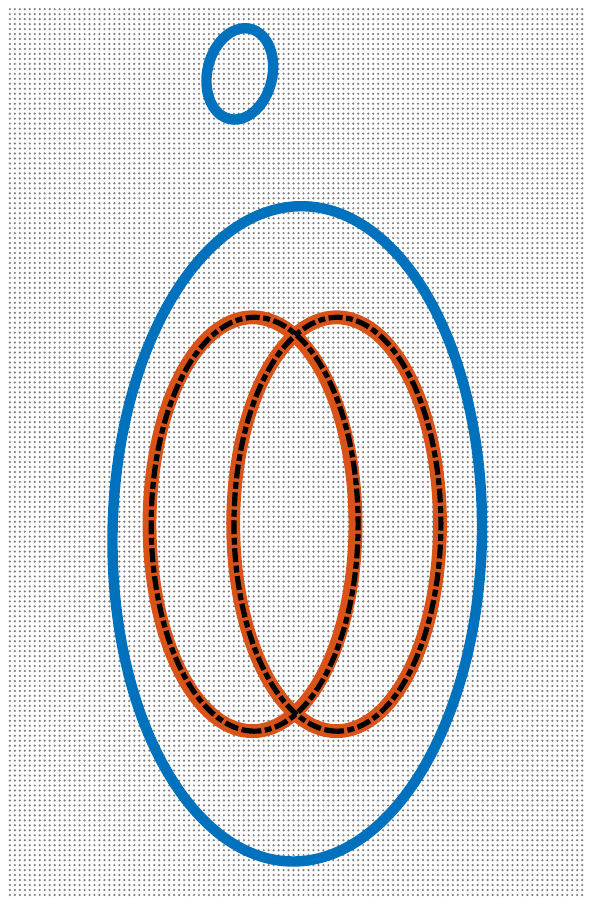}} 
    \end{center} 
    \caption{\label{fig:eyecatcher} Two-dimensional \ac{GM} parameter estimation using EM from \cite{Gebru16} (blue line), and EM according to our proposed method (red line). Compare the ground truth (black line). Equidistant samples (grey dots) were weighted with the \ac{GM} density function and given to the EM algorithms. } 
    % Generating Code: ISAS/Projects/EMFilter/matlab/EMEval03.m
    % Figures: ISAS/Projects/EMFilter/matlab/figures/EMEval03/
\end{figure}

%
% Surprise
%
So it comes as a surprise that \ac{GM} estimation for \emph{weighted samples} is hard to find in literature.
%
% Standard reference: wrong
%
It might be even more surprising that the standard reference \cite{Gebru16} gives incorrect results, see \Fig{fig:eyecatcher}.

\section{Context} 
%
% Applications
%
Applications for sample-to-density function approximation include clustering of unlabled data \cite{jain1999,larsen2002}, multi-target tracking \cite{Prembida06,Clark2007}, group tracking \cite{Pang2011}, multilateration \cite{Tzoreff2017,Li2018}, and arbitrary density representation in nonlinear filters \cite{Dovera2011,FUSION20_Frisch}. 

%
% k-means
%
A popular basic solution to this is the $k$-means algorithm. It does not find a complete density representation, only the means of the individual clusters. The $k$-means algorithm uses hard sample-to-mean associations, therefore yields merely approximate solutions but can be computationally optimized using $k$-d trees \cite{Kanungo2002,Hamerly2002}. Moreover, the global optimum can be found deterministically \cite{Likas2003}, therefore it can be used to provide an initial guess for more elaborate algorithms. 

%
% ML Newton
%
A sample-to-density approximation that is optimal in a maximum likelihood sense can be searched with numerical optimization techniques such as the Newton algorithm that has quadratic convergence but high computational demand per iteration, quasi-Newton methods, the method of scoring, or the conjugate gradient method with slower convergence but less computational effort per iteration \cite{Redner84_EM}. 

\subsection{State-of-the-art} 
%
% EM : Pros and Cons
%
The \ac{EM} algorithm has been used for decades \cite{Dempster77,wu83_EM} to solve statistical problems. It converges rather slowly, especially if the \ac{GM} components are poorly separated, but it provides a valid parameter set in every iteration step, i.e., nonnegative and normalized component weights and positive semidefinite covariance matrices, without the need of any artificial safeguards. The \ac{EM} algorithm features good global convergence to some local optimum, is very easy to implement, has low computational cost per iteration when using optimized libraries for standard statistical tasks, and needs little storage \cite{Redner84_EM}. 
%
% No. of Components
%
There are extensions of the \ac{EM} to automatically determine the optimal number of Gaussian components \cite{Figueiredo2002,Zhang2004,Melynkov2012}. 

\subsection{Contribution} 

The contribution of this paper is a fast, simple, and practical \ac{EM} method for the correct treatment of weighted samples in Gaussian mixture estimation.

\section{Problem Formulation} 
%
% Samples
%
For an observed set of $L$ weighted samples 
\begin{equation*}
    \obsrv = \braces{ \braces{\ws_1, \s_1}, \braces{\ws_2, \s_2}, \hdots, \braces{\ws_L, \s_L} } 
\end{equation*}
with sample locations $\s_i$ as vectors in the $D$-dimensional Euclidean space $\NewR^D$, and scalar weights $\ws_i$, 
we want to find a \ac{GM} density function with $M$ Gaussian components
%
% GM 
%
\begin{align}
    f(\x|\param) &= \sum_{m=1}^M \w_m \, \Gauss\!\paren{\x-\mean_m,\;\C_m} \enspace, \\
    \Gauss\!\paren{\x-\mean,\;\C} &= \frac{\exp\!\braces{ -\frac{1}{2} \paren{\x-\mean}\T\! \C^{-1} \paren{\x-\mean} } }{\sqrt{\abs{2\pi\C}}} \enspace,
\end{align}
with nonnegative component weights $\w_m\geq 0$ that are normalized, $\sum_{m=1}^M \w_m = 1$, component means $\mean_m \in \NewR^D$, and component covariances $\C_m \in \NewR^{D \times D}$. The \ac{GM} should explain the observed samples as good as possible. 
%
% ML 
%
We thus estimate \ac{GM} parameters $\param$ 
\begin{align}
    \param &= \braces{ \braces{\w_1,\mean_1,\C_1},\; \hdots,\; \braces{\w_M,\mean_M,\C_M} } 
\end{align}
from the weighted samples $\obsrv$, ideally in a maximum likelihood sense
\begin{align}
	\widehat\param^{\op{ML}} &= \argmax_\param\braces{ \fd{\obsrv | \param}{}(\obsrv \,|\, \param) } \enspace.
\end{align}

%
% Possible methods
%
This can be done via numerical optimization or, more efficiently, using the \ac{EM} algorithm. 
%
% Hidden assignment
%
For the \ac{EM} algorithm, we additionally consider a hidden variable $\hidden$. It contains the association probabilities $\eta_{i,m}$ between samples $\braces{\ws_i, \s_i}$ and \ac{GM} components $\braces{\w_m,\mean_m,\C_m}\,$. 

% [Note: differing variable names in Gebru \cite{Gebru16}: parameters $\mat \Theta$, observed data $\mat X$, hidden (assignment variables) $\mat Z$.]

\section{Key Idea}
%
% Gedankenexperiment
%
We believe that the following two things should give the same contribution to the result: First, one sample with double weight, and second, two single-weight samples that are arranged with infinitesimally small or zero distance. 
%
% No Covariance Rescaling
%
Therefore, we propose to determine the hidden association parameters $\hidden$ only based on sample locations. In other words we use the observed sample weights only in the maximization step and not in the expectation step. 

%
% Maximization
%
For the maximization step, we propose to estimate \ac{GM} component weights, means, and covariances as a weighted average, where weighting is the product of observed sample weights and sample-to-\ac{GM} component associations. 

% Normalization of sample weights does not matter as we do not use the samples in the exponential function. 

\section{Implementation of Proposed Method}
\label{Sec:Implementation}
Associations $\hidden$ between Samples and \ac{GM} components are unknown but necessary for an \ac{EM} algorithm in order to independently calculate moments of individual mixtures. Marginalization over all possible associations 
\begin{align}
	\fd{\obsrv | \param}{}(\obsrv | \param) &= \int \fd{\hidden,\obsrv | \param}{}(\hidden,\obsrv \,|\, \param) \dd \hidden \enspace, \label{eq:EM} 
\end{align}
is infeasible, hence the separation into expectation and maximization steps according to the \ac{EM} algorithm.

\subsection{Expectation Step}
\label{Sec:Implementation:Expectation}
Besides the given observed data $\obsrv$, we assume an estimate $\paramStep{r}$ of the parameter vector containing the \ac{GM} parameters $\braces{\w_m^{(r)},\mean_m^{(r)},\C_m^{(r)}}\,$,\; $m\in \braces{1,\hdots,M}$, with iteration index $(r)$, to obtain a new estimate of the hidden data $\hiddenStep{r+1}$
\begin{align}
    \eta_{i,m}^{(r+1)}
    = \frac{\w_m \,  \Gauss\!\paren{\s_i-\mean_m^{(r)},\;\C_m^{(r)}}} 
    {\sum_{\widetilde m=1}^{M} \w_{\widetilde m} \,  \Gauss\!\paren{\s_i-\mean_{\widetilde m}^{(r)},\;\C_{\widetilde m}^{(r)}}} \enspace, \label{eq:assoc} 
\end{align}
with matrix elements $\brackets{\hiddenStep{r+1}}_{i,m}\!=\eta_{i,m}^{(r+1)}$. Due to the normalization such that the row sum is equal to one, $\hiddenStep{r+1}$ describes a ``probability of association'' for each sample $i$ to each component $m$ of the \ac{GM}.

\subsection{Maximization Step} 
\label{Sec:Implementation:Maximization}
Using said estimate of the hidden data $\hiddenStep{r+1}$ and also the observed data $\obsrv$, i.e., sample locations $\s_i$ and sample weights $\ws_i$, we obtain a new estimate of the parameter vector $\paramStep{r+1}$
%
%\begin{align}
%	\paramStep{r+1} &= \argmax_\param\braces{\fd{\hidden,\obsrv | \param}{}(\hiddenStep{r},\obsrv \,|\, \param) }\enspace.
%\end{align}
%
\begin{align}
    \w_m^{(r+1)} &= \frac{ \sum_{i=1}^L \eta_{i,m}^{(r+1)}  \ws_i }
    {\sum_{\widetilde m=1}^M \sum_{\tilde i=1}^L \eta_{\tilde i,\widetilde m}^{(r+1)}  \ws_{\tilde i} }  \label{eq:comp-weight} \enspace, 
    \\ 
    \mean_m^{(r+1)} &= \frac{ \sum_{i=1}^L \eta_{i,m}^{(r+1)}  \ws_i  \, \s_i }
    {\sum_{\tilde i=1}^L \eta_{\tilde i,m}^{(r+1)} \ws_{\tilde i}} \label{eq:comp-mean} \enspace, 
    \\ 
    \C_m^{(r+1)} &= \frac{ \sum_{i=1}^L \eta_{i,m}^{(r+1)}  \ws_i  \paren{\s_i-\mean_m^{(r+1)}}\paren{\s_i-\mean_m^{(r+1)}}\T }
    {\sum_{\tilde i=1}^L \eta_{\tilde i,m}^{(r+1)} \ws_{\tilde i}}  \label{eq:comp-cov} \enspace.  
\end{align}

\begin{figure*}[!ht] 
    \begin{center} 
        \subfloat[Weights; ``separated'' components] {\includegraphics[width=.33\linewidth] {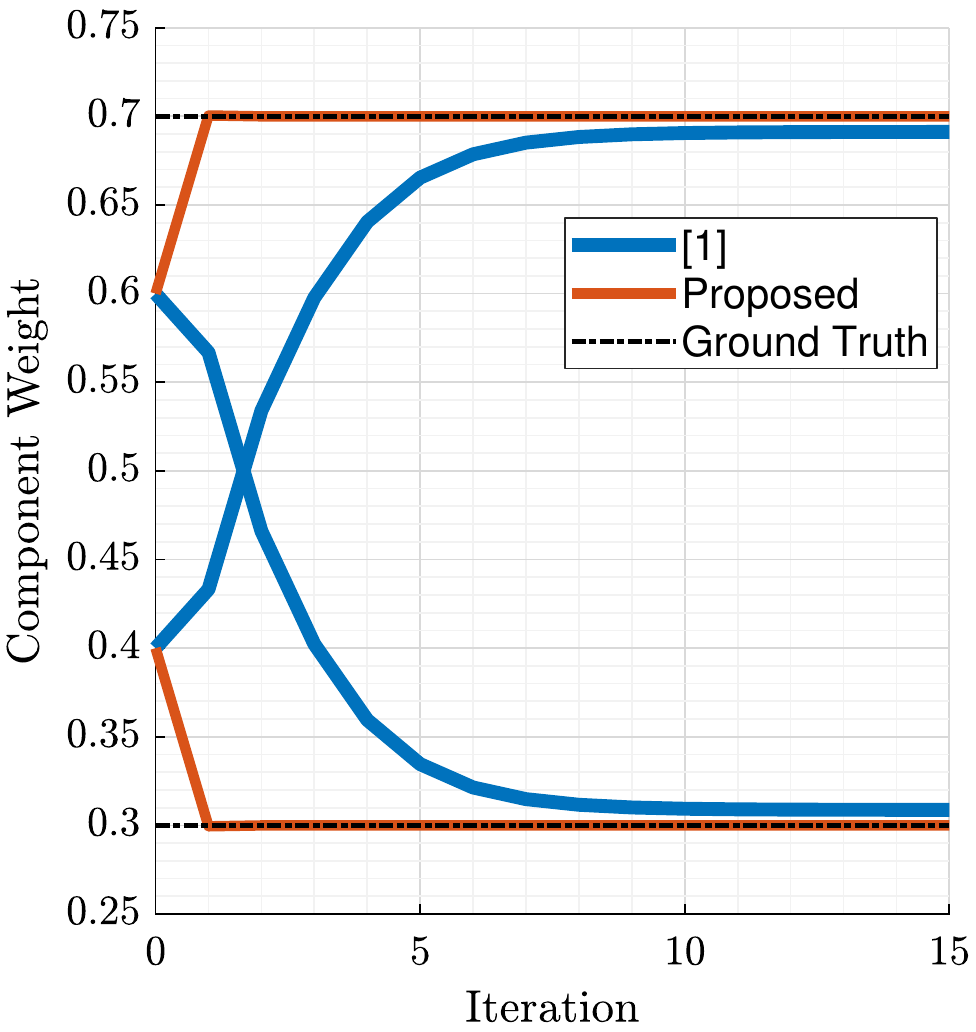}} 
        \subfloat[Means; ``separated'' components] {\includegraphics[width=.33\linewidth] {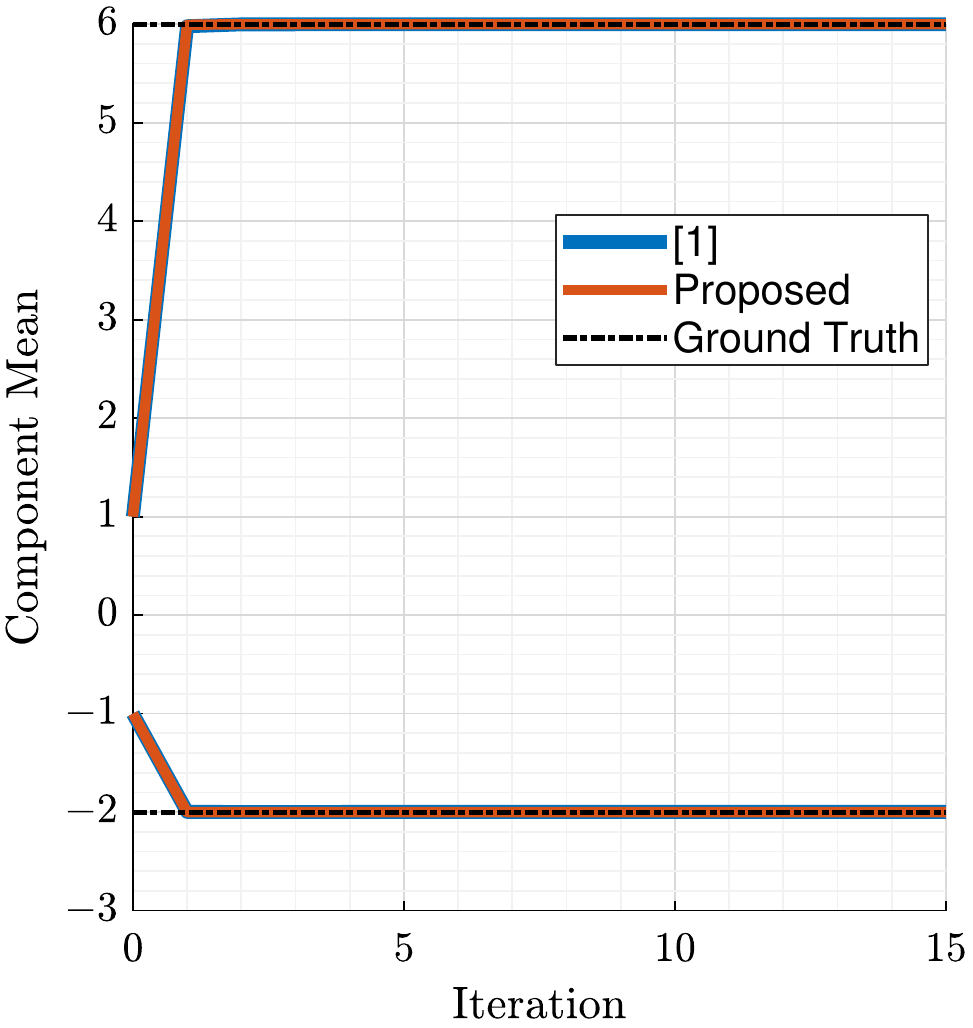}} 
        \subfloat[Standard deviations; ``separated'' components] {\includegraphics[width=.33\linewidth] {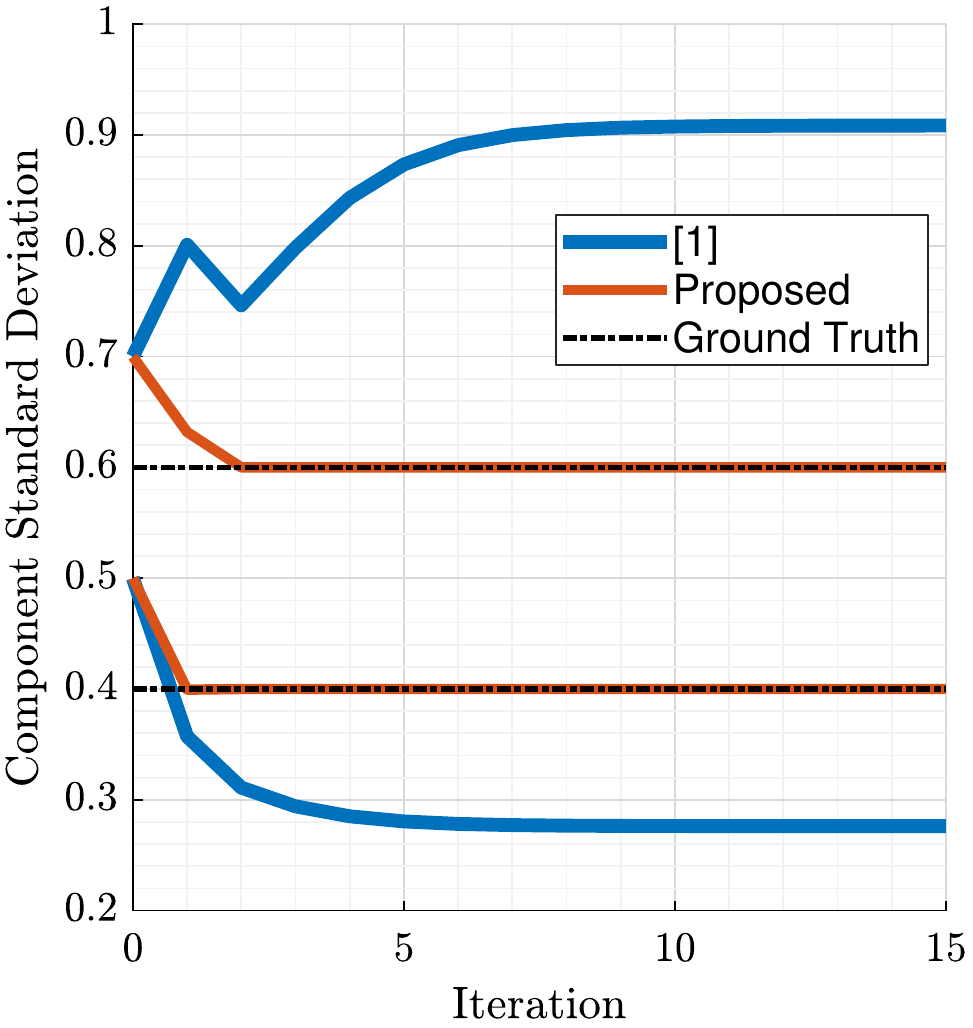}} \\
        \subfloat[Weights; ``overlapping'' components] {\includegraphics[width=.33\linewidth] {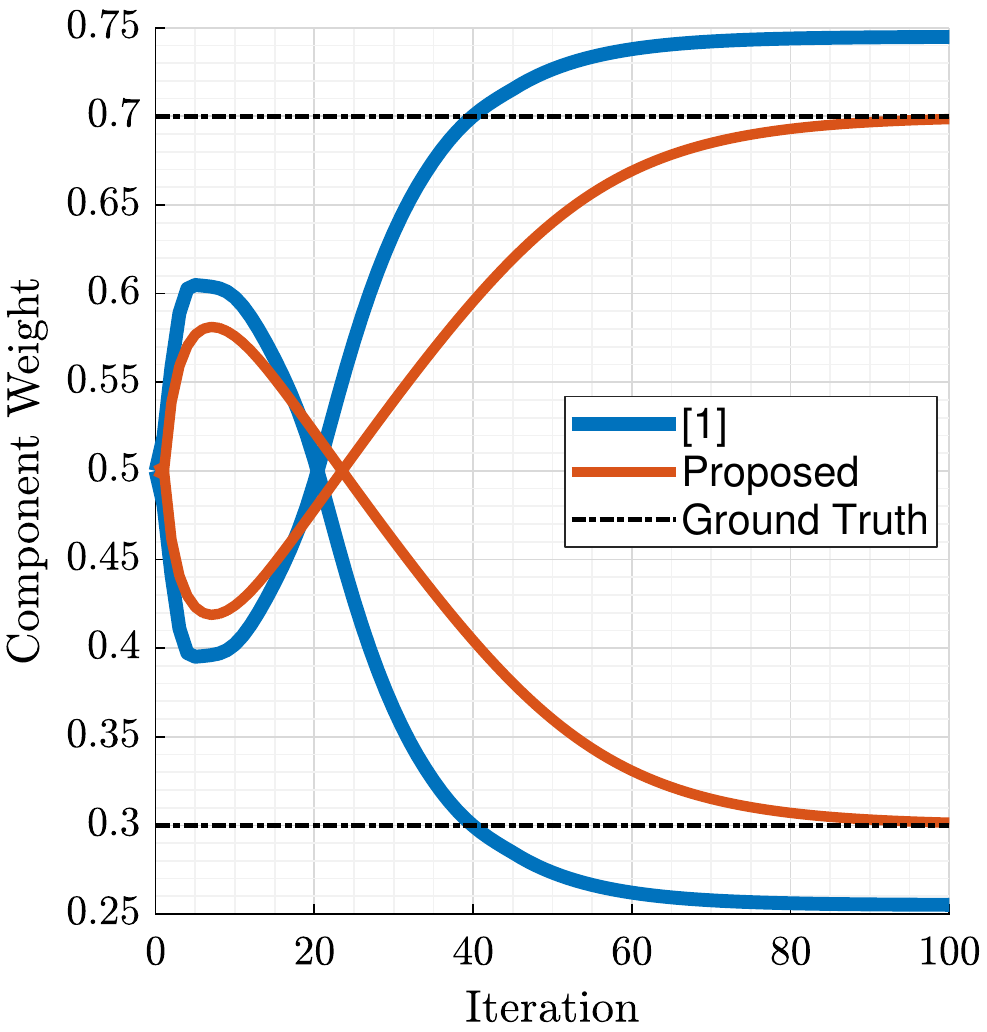}} 
        \subfloat[Means; ``overlapping'' components] {\includegraphics[width=.33\linewidth] {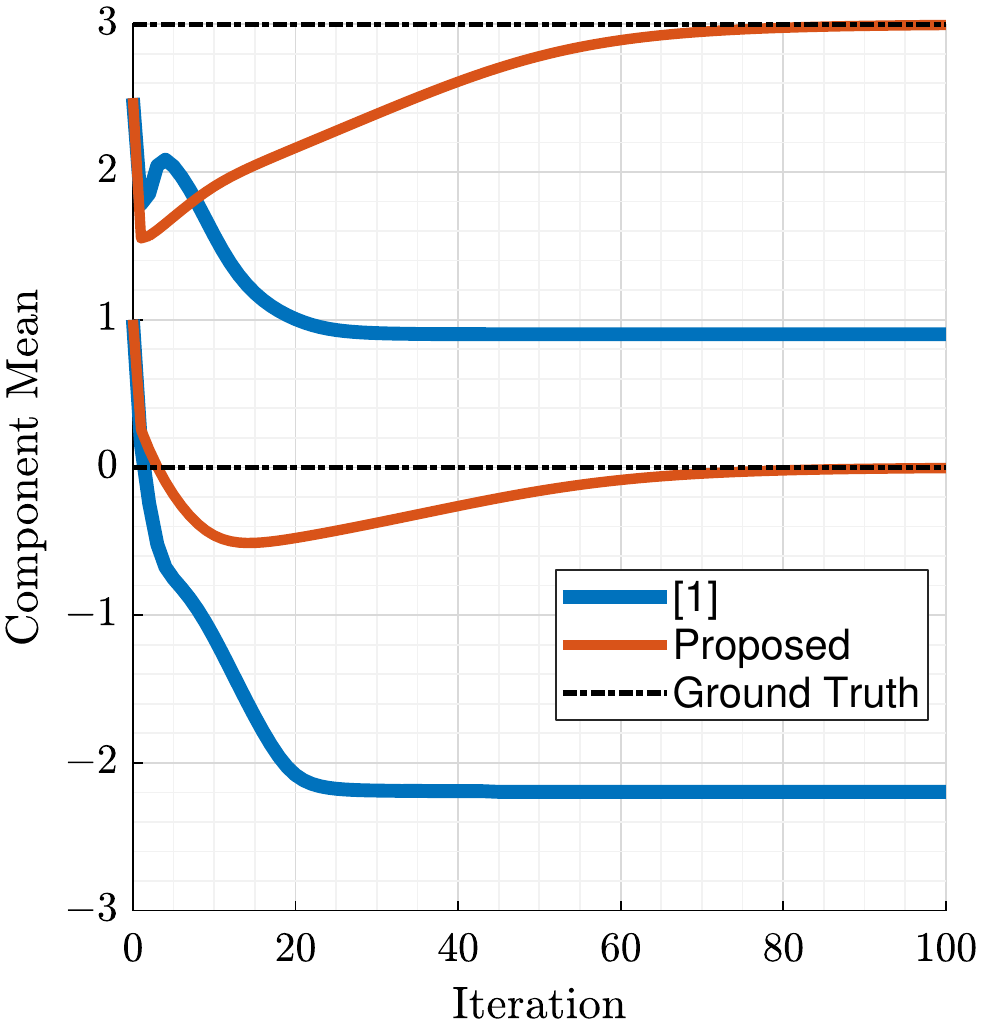}} 
        \subfloat[Standard deviations; ``overlapping'' components] {\includegraphics[width=.33\linewidth] {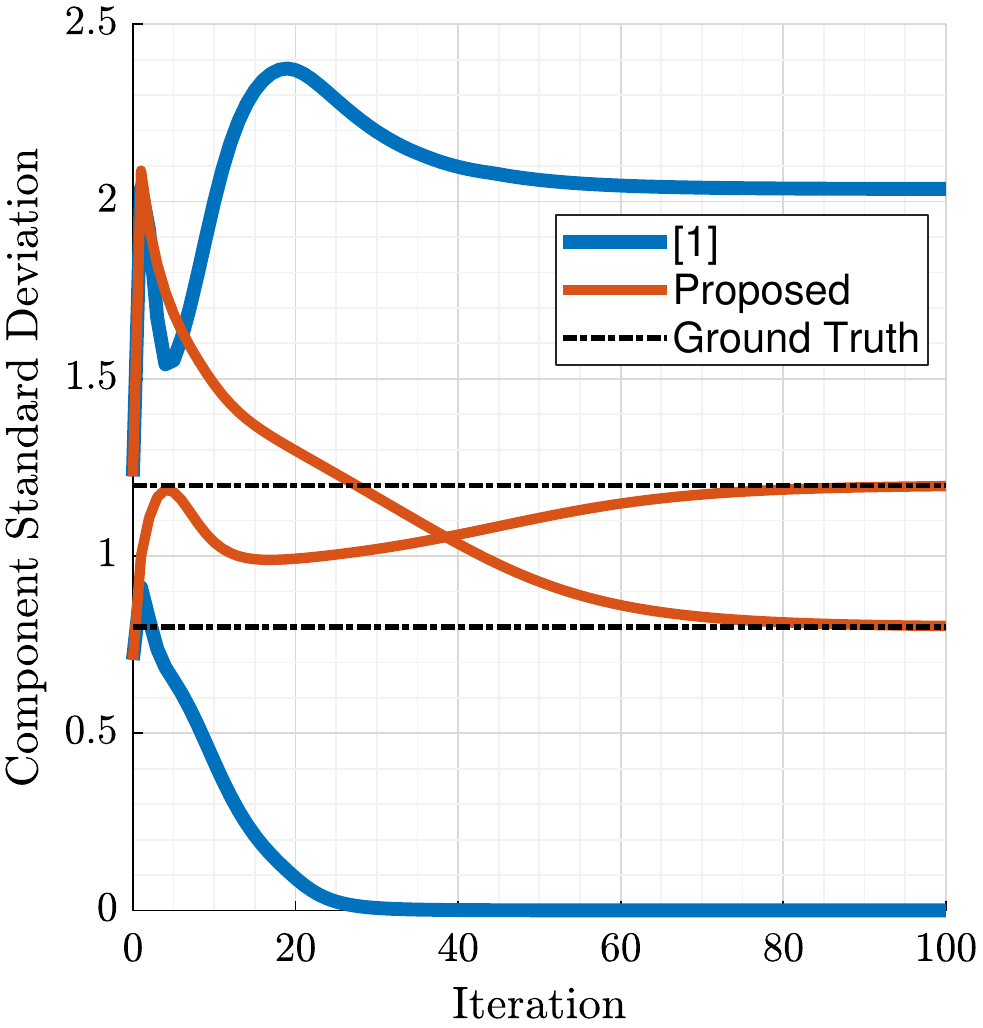}} \\
        \subfloat[Densities; ``separated'' components] {\includegraphics[width=.49\linewidth] {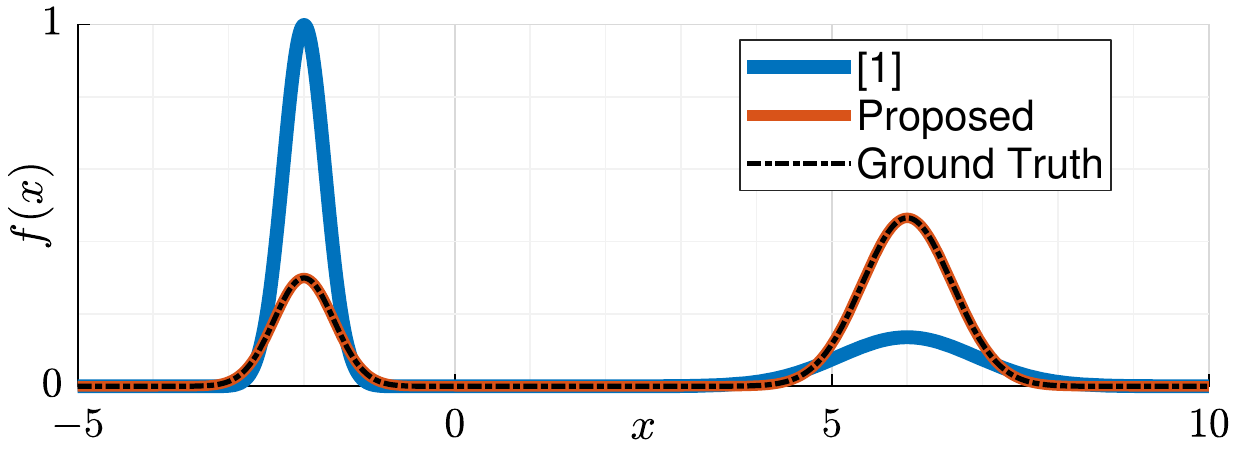}} 
        \subfloat[Densities; ``overlapping'' components] {\includegraphics[width=.49\linewidth] {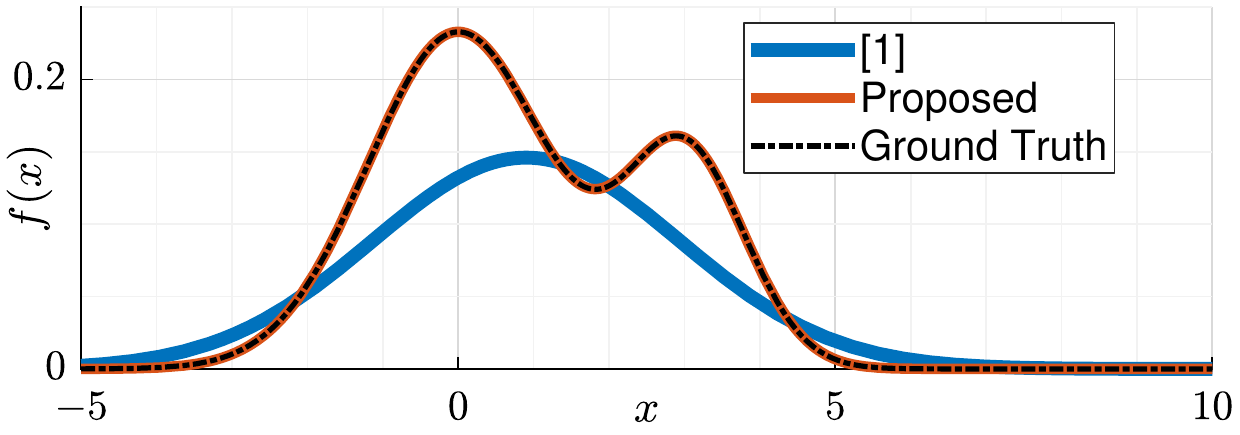}} 
    \end{center} 
    \caption{\label{fig:SimpleEval} A simple scalar example with two \ac{GM} components. Equidistant samples were weighted with the ground truth probability density function, and the \ac{GM} parameters (component weights, means, and variances) were estimated with our proposed method (red lines) and the method from \cite{Gebru16} (blue lines). Ideally, the estimations should converge to the ground truth (black dashed lines) after some iterations. } 
    % Generating Code: ISAS/Projects/EMFilter/matlab/EMEval02.m 
    % Figures: ISAS/Projects/EMFilter/matlab/figures/EMEval02/
\end{figure*}

\subsection{Split Sample Linearity}
The obtained parameter estimate $\paramStep{r+1}$, after performing one expectation and maximization step for some given prior parameter estimate $\paramStep{r}$, is identical whether we have a set of weighted samples
\begin{align*}
    \obsrv = \braces{ \braces{\ws_1, \s_1}, \braces{\ws_2, \s_2}, \hdots, \braces{\ws_L, \s_L} } \;,
\end{align*}
or ``split samples'' with, e.g., two samples and two weights at each sample location, i.e., 
\begin{align}
    \widetilde \obsrv =& \bigg\{ \braces{\ws_1^{(1)}\!,\, \s_1},  \braces{\ws_1^{(2)}\!,\, \s_1},\; 
    \braces{\ws_2^{(1)}\!,\, \s_2}, \braces{\ws_2^{(2)}\!,\, \s_2}, \\
    & \quad \hdots,\; \braces{\ws_L^{(1)}\!,\, \s_L}, \braces{\ws_L^{(2)}\!,\,\s_L} \bigg\} \enspace, \label{eq:split-samples}
\end{align}
with $\ws_i=\ws_i^{(1)}+\ws_i^{(2)} \; \forall \; i \in \braces{1,\hdots,L}$. This is because association probabilities $\eta_{i,m}^{(r+1)}$ in the expectation step do not depend on sample weights $\ws_i$, and for the maximization step due to its linearity it does not matter whether there are two samples with weights $\ws_i^{(1)}\!,\, \ws_i^{(2)}$ at the same location $\s_i\,$, or only one sample that contains their combined weight $\ws_i$.

%
% More general 
%
Note that the same holds for any other linear combination of more than two weights and samples at each same sample location, moreover not all but only a few sample locations may exhibit ``split samples''. We see this invariance against ``split samples'' as a logical sanity check the method should pass in order to be consistent. 

%\newpage
\section{Implementation in \cite{Gebru16}}
\label{Sec:Implementation-Gebru}
% Code:    https://github.com/isrish/EM_WD
% Project: https://team.inria.fr/perception/research/wdgmm/
For comparison, we quote the implementation from \cite{Gebru16,Gebru14} and highlight the differences to what we propose. 

\subsection{Expectation Step in \cite{Gebru16}}
\label{Sec:Implementation-Gebru:Expectation}
%
% Association
%
For estimating the associations $\eta_{i,m}^{(r+1)}$ between samples $\s_i$ and \ac{GM} components $\braces{\w_m^{(r)},\mean_m^{(r)},\C_m^{(r)}}$, the covariances $\C_m^{(r)}$ of the individual Gaussian components are scaled in \cite{Gebru16} based on the sample weights $\ws_i$
\begin{align}
    %&\brackets{\hiddenStep{r+1}}_{i,m} =
    \eta_{i,m}^{(r+1)} = 
    \frac{ \w_m \, \Gauss\!\paren{\s_i-\mean_m^{(r)},\; \C_m^{(r)}/\ws_i} }
    { \sum_{\widetilde m=1}^M \w_{\widetilde m} \, \Gauss\!\paren{\s_i-\mean_{\widetilde m}^{(r)},\; \C_{\widetilde m}^{(r)}/\ws_i} } \enspace.
    % https://github.com/isrish/EM_WD/blob/master/EM_WDF.m  -  L.119
    % https://arxiv.org/pdf/1509.01509.pdf  -  (8) 
\end{align}
We however propose to use the \ac{GM} covariances $\C_m^{(r)}$ without any sample-specific adaptions \eqref{eq:assoc}. 

\subsection{Maximization Step in \cite{Gebru16}} 
\label{Sec:Implementation-Gebru:Maximization}
%
% Component Weights
%
In \cite{Gebru16}, sample weights $\ws_i$ are not considered when calculating the Gaussian mixture component weights $\w_m^{(r+1)}$  
\begin{align}
    \w_m^{(r+1)} &= \frac{1}{L} \sum_{i=1}^L \eta_{i,m}^{(r+1)} \enspace,
    % https://github.com/isrish/EM_WD/blob/master/EM_WDF.m  -  L.153
    % https://arxiv.org/pdf/1509.01509.pdf  -  (8) 
\end{align}
compare \eqref{eq:comp-weight}. 
%
% Component Weights
%
Calculation of component means $\mean_m^{(r+1)}$  
\begin{align}
    \mean_m^{(r+1)} &= \frac{ \sum_{i=1}^L \eta_{i,m}^{(r+1)}  \ws_i  \s_i  }
    {\sum_{\tilde i=1}^L \eta_{\tilde i,m}^{(r+1)}  \ws_{\tilde i} } 
    % https://github.com/isrish/EM_WD/blob/master/EM_WDF.m  -  L.141
    % https://arxiv.org/pdf/1509.01509.pdf  -  (9) 
\end{align}
is identical to our proposed method \eqref{eq:comp-mean}. 
%
% Difference
%
For component covariance estimation $\C_m^{(r+1)}$, the difference between \cite{Gebru16} and our proposed method \eqref{eq:comp-cov} is that sample weights $\ws_i$ are not considered for normalization 
\begin{align}
    \C_m^{(r+1)} &= \frac{ \sum_{i=1}^L \eta_{i,m}^{(r+1)} \ws_i  \paren{\s_i-\mean_m^{(r+1)}}\paren{\s_i-\mean_m^{(r+1)}}\T }
    {\sum_{\tilde i=1}^L \eta_{\tilde i,m}^{(r+1)}}  \enspace. 
    % https://github.com/isrish/EM_WD/blob/master/EM_WDF.m  -  L.150
    % https://arxiv.org/pdf/1509.01509.pdf  -  (10) 
\end{align}

\subsection{Split Sample Linearity in \cite{Gebru16}}
For the \ac{EM} method according to \cite{Gebru16}, the result of each iteration is different when we ``split'' some samples in different ways, e.g., in two parts \eqref{eq:split-samples}. Therefore, a double sample weight is \emph{not} equivalent with two samples at the same location. The evaluation section will demonstrate that not only the individual iteration results, but also the final result differs from our proposed method, and from the ground truth. 

\section{Evaluation and Comparison with \cite{Gebru16}}

%
% MWE 1D, M=2
%
As the simplest example, we define a one-dimensional \ac{GM} with two components. 
%
% Weighted Grid
%
A large number of equidistant samples is placed in the relevant region, and the \ac{GM} density function at each sample location is used as the respective sample weight. 
%
% Initial Guess
%
Furthermore, some random initial guess of the \ac{GM} parameters is given. 
%
% Comparison
%
Two algorithms are compared in solving this density estimation problem. First, our proposed method as defined in \Sec{Sec:Implementation}, and second, the method as proposed in \cite{Gebru16} and replicated here in \Sec{Sec:Implementation-Gebru}. 
%And third, a combination of both methods, with the expectation step from \Sec{Sec:Implementation-Gebru:Expectation}, but the maximization step according to \Sec{Sec:Implementation:Maximization}. 

%
% "Separated" 
%
One setup is defined where the two Gaussian components are rather ``separated'', this can be solved with about 15 iteration steps, see \Fig{fig:SimpleEval} (a, b, c). 
%
% "Overlapping"
%
A second setup has Gaussian components that are closer together and exhibit some ``overlap'' of probability mass. Both \ac{EM} algorithms need much more iteration steps to converge here, see \Fig{fig:SimpleEval} (d, e, f). 

%
% Interpretation "Separated"
%
For the ``separated'' Gaussian components we find that all algorithms provide a very good estimation of the \ac{GM} component means after about three iterations. The weighting factor estimates need more iterations to converge and are slightly off with the algorithm from \cite{Gebru16}. Standard deviations from \cite{Gebru16} are not reliable at all. Only our proposed method provides accurate results here. 
%
% Interpretation "Overlapping"
%
In the ``overlapping'' setup, the \ac{GM} component weight, mean, and variance estimates converge to solutions that are significantly off the ground truth when using the method from \cite{Gebru16}. Our proposed method needs more iterations to converge but finds the accurate solution in the end. 

\section{Conclusions} 
% Wofür benutzen, z.B. Speaker Localization % Re-Weighting Filtering 
% ML Clustering Unsupervised Learning
% K-Means (weighted available?) 

%
% New applications
%
Considering weighted samples opens new applications for \ac{GM} estimation, e.g., in the field of Bayesian estimation, see \cite{FUSION20_Frisch}.
%
% Correct treatment
%
The correct treatment of weighted samples in \ac{GM} estimation was derived.
%
% Literature: wrong
%
It was shown that current approaches have a serious flaw that leads to wrong estimates.
%
% Simple modification of existing code
%
The proper modifications can simply be added to existing \ac{GM} estimation code to extend its applicability to weighted samples.
%
% Plugin replacement, backwards compatible
%
The proposed method is also a plugin replacement for standard \ac{GM} estimators as it is backwards compatible for unweighted samples.

%\clearpage
\bibliographystyle{bib/IEEEtran}
\bibliography{bib/IEEEabrv, bib/ISASPublikationen, bib/paperlocal}

\end{document}